\documentclass[letterpaper, 10 pt, conference]{ieeeconf}  % Comment this line out if you need a4paper

\IEEEoverridecommandlockouts                              % This command is only needed if 
\usepackage{mathptmx} % assumes new font selection scheme installed
\usepackage{times} % assumes new font selection scheme installed
\usepackage{amsmath} % assumes amsmath package installed
\usepackage{amssymb}  % assumes amsmath package installed
\usepackage{color,soul}

\title{\LARGE \bf
Autonomous Control for Orographic Soaring of Fixed-Wing UAVs
}

\author{Tom Suys$^{1}$, Sunyou Hwang$^{1}$, Guido C.H.E. de Croon$^{1}$, and Bart D.W. Remes$^{1}$% <-this % stops a space
\thanks{$^{1}$All authors are with the MAVLab, Department of Control and Operations, Faculty of Aerospace Engineering, Delft University of Technology, 2629HS Delft, the Netherlands
         {\tt\small mail@tomsuys.com, S.Hwang-1@tudelft.nl, G.C.H.E.deCroon@tudelft.nl, B.D.W.Remes@tudelft.nl}}%
\thanks{A supplementary video of the flight tests is available at: \url{https://youtu.be/b_YLoinHepo}}%    
}

\usepackage{cite}
\usepackage{algorithmic}
\usepackage{graphicx}
\usepackage{textcomp}
\usepackage{xcolor}
\usepackage{hyperref}
\usepackage{etoolbox}

\begin{document}
\providetoggle{changes}
\settoggle{changes}{false}

\maketitle
\thispagestyle{empty}
\pagestyle{empty}

%%%%%%%%%%%%%%%%%%%%%%%%%%%%%%%%%%%%%%%%%%%%%%%%%%%%%%%%%%%%%%%%%%%%%%%%%%%%%%%%
%%%%%%%%%%%%%%%%%%%%%%%%%%%%%%%%%%%%%%%%%%%%%%%%%%%%%%%%%%%%%%%%%%%%%%%%%%%%%%%%
%%%%%%%%%%%%%%%%%%%%%%%%%%%%%%%%%%%%%%%%%%%%%%%%%%%%%%%%%%%%%%%%%%%%%%%%%%%%%%%%

\begin{abstract}
We present a novel controller for fixed-wing UAVs that enables autonomous soaring in an orographic wind field, extending flight endurance. Our method identifies soaring regions and addresses position control challenges by introducing a target gradient line (TGL) on which the UAV achieves an equilibrium soaring position, where sink rate and updraft are balanced. Experimental testing validates the controller's effectiveness in maintaining autonomous soaring flight without using any thrust in a non-static wind field. We also demonstrate a single degree of control freedom in a soaring position through manipulation of the TGL.

% We present a novel controller for fixed-wing UAVs that enables autonomous soaring in an orographic wind field, extending flight endurance. Our method identifies feasible soaring regions and addresses position control challenges by introducing a target gradient line on which it achieves equilibrium soaring position where sink rate and updraft are balanced. Experimental testing validates the controller's effectiveness in maintaining autonomous soaring flight in a non-static wind field. We also demonstrate a single degree of control freedom in the soaring position through manipulation of the target gradient line.

\end{abstract}

% %vvvvvvvvvvvvvvvvvvvvvvvvvvvvvvvvvvv
% \iftoggle{changes}{\hl{
% ORIGINAL - CONCISENESS - TECHNICAL LANGUAGE\\
% Flight endurance often limits the application and potential in Unmanned Aerial Vehicle (UAV) missions. We propose a novel approach to extend the endurance of fixed-wing UAVs by presenting a controller capable of soaring in an orographic wind field. This research presents a model to highlight feasible soaring regions and address the difficulty in maintaining position control. The concept of a target gradient line is introduced as part of the control algorithm that autonomously finds the equilibrium soaring position where sink rate and updraft are balanced. An experimental test setup was constructed, which validated the control algorithm. The test data shows promising effectiveness in the controller to maintain autonomous soaring flight in a non-static wind field. Furthermore, a single degree of control freedom in the soaring position can be realized by manipulating the target gradient line. 
% }}
% %^^^^^^^^^^^^^^^^^^^^^^^^^^^^^^^^^^^

\begin{keywords}
wind hovering, orographic soaring, autonomous control, UAV
\end{keywords}

%%%%%%%%%%%%%%%%%%%%%%%%%%%%%%%%%%%%%%%%%%%%%%%%%%%%%%%%%%%%%%%%%%%%%%%%%%%%%%%%
%%%%%%%%%%%%%%%%%%%%%%%%%%%%%%%%%%%%%%%%%%%%%%%%%%%%%%%%%%%%%%%%%%%%%%%%%%%%%%%%
%%%%%%%%%%%%%%%%%%%%%%%%%%%%%%%%%%%%%%%%%%%%%%%%%%%%%%%%%%%%%%%%%%%%%%%%%%%%%%%%

\section{INTRODUCTION}

UAVs have benefited from advancements in battery technology and miniaturization of avionics, which resulted in an increase in their endurance and range. However, the full potential of UAV applications remains limited by reduced flight time. Therefore, it is useful to research other techniques that can positively impact the effective endurance of these UAVs. An interesting technique that has shown great potential is exploiting updrafts to stay airborne nearly indefinitely. Albatrosses have perfected this technique of soaring \cite{Rayleigh1, Rayleigh2}, allowing them to fly without mechanical energy cost and embark on journeys exceeding 20,000 km, staying airborne for multiple days at a time \cite{Alb_end, Alb_eng}. 

All soaring techniques aim to extract sufficient energy to stay airborne without losing altitude. Dynamic soaring is when energy is extracted due to a gradient in the horizontal wind velocity \cite{Zhao}, while static soaring relies on a vertical, upward wind component. Static soaring includes two types: thermal soaring, which is created by a rising column of air, and orographic soaring, which is created by the upwards deflection of the wind stream. This research focuses exclusively on orographic soaring. 

\begin{figure}[ht]
\centerline{\includegraphics[width=1\linewidth]{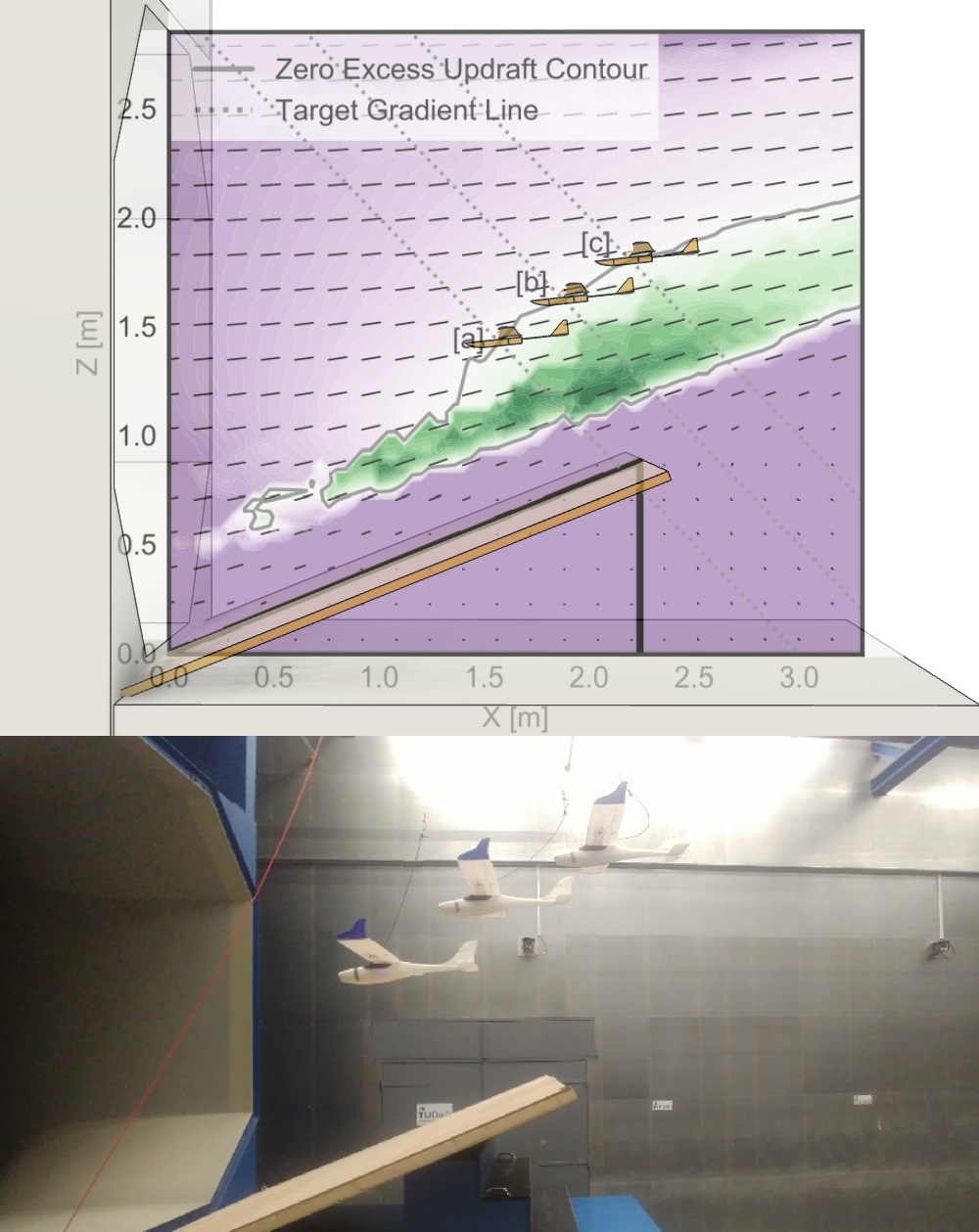}}
\caption{Autonomous soaring and position control of an UAV. The model shows the predicted excess updraft in the orographic wind field of the test setup. By precisely maintaining position control along an operator-specified target gradient line, the UAV successfully achieves autonomous soaring flight at the intersection with zero-excess updraft. The observed flight data, presented as three stacked stills in the image, correspond to the predicted soaring locations for three distinct target gradient lines specified in the test. \vspace{-10mm}}
\label{intropic}
\end{figure}

% Prior research has explored the potential of exploiting orographic updrafts \cite{White, White2, Mohamed, Mohamed2, Guerra-Langan, Langelaan} . White et al. utilized simulations and measurement data to evaluate the feasibility of soaring in the updraft generated by tall buildings. However, flight demonstrations were not conducted in these studies. While some studies have demonstrated orographic soaring \cite{Fisher, DeJong},they require a priori knowledge of the wind field and offer limited control freedom. 

Prior research has explored the potential of exploiting orographic updrafts \cite{White, White2, Mohamed, Mohamed2, Guerra-Langan, Langelaan}. White et al. utilized simulations and measurement data to evaluate the feasibility of soaring in the updraft generated by tall buildings. However, flight demonstrations were not conducted in these studies. While some studies have demonstrated orographic soaring \cite{Fisher, DeJong}, they require a priori knowledge of the wind field and manual control of a human pilot to steer the UAV to an initial soaring position. Above studies considered a static wind field and predetermined path of the UAV.

In contrast, with this paper we propose a novel orographic soaring method with a single degree of control freedom, which can adapt to a non-static wind field, and does not require any throttle usage. Our method utilizes information derived from the approximate location of the updraft core but does not require complete prior knowledge of the wind field. We demonstrate fully autonomous soaring flight, and a single degree of position control of a UAV without propeller in a non-static wind field in a real-world test.
We derive the kinematics involved with orographic soaring in \autoref{oro_soa}. Its distinct control freedom is outlined in \autoref{soa_con_fre}. Potential flow is used to estimate a soaring wind field in \autoref{win_ana}. In \autoref{aut_con_str}, a control strategy is proposed to maintain autonomous soaring flight. The experimental test setup and test results are discussed in \autoref{exp_tes_set} and \autoref{Test_res}. % respectively. 

\section{OROGRAPHIC SOARING}\label{oro_soa}
The kinematics involved with orographic soaring can be modeled using a point mass model, as illustrated by Langelaan \cite{langelaanpointmass}. In this model, we define the vehicle mass ($m$), angle of attack ($\alpha$), thrust ($T$), drag ($D$), and lift force ($L$) as key parameters. 

First, considering the forces parallel and perpendicular to the flight path, we obtain two equations:
\begin{align}
% \begin{equation}
mg \cos(\gamma) = L + T \sin(\alpha) \\
% \end{equation}
% \begin{equation}
\label{eq:drag}
mg \sin(\gamma) = D - T \cos(\alpha)
% \end{equation}
\end{align}

Here, flight path angle $\gamma$ is assumed small and a small angle approximation is used. Next, we obtain an equation that relates lift force to other parameters:
\begin{equation}
mg = L = \frac{1}{2}\rho v_a^2 S C_L
\end{equation}

Using this equation, we can calculate the lift coefficient ($C_L$) in terms of other parameters:
\begin{equation}
C_L = \frac{2mg}{\rho v_a^2 S}
\end{equation}

Additionally, we can derive a second-order approximation for the drag force ($D$):
\begin{equation}
D = \frac{1}{2} \rho v_a^2 S (a_0 + a_1 C_L + a_2 C_L^2)
\end{equation}

Substituting this approximation into Equation \ref{eq:drag}, we can calculate the flight path angle for a given speed and thrust:
\begin{equation}
m g \gamma = \frac{1}{2} \rho v_a^2 (a_0 + a_1 C_L + a_2 C_L^2) - T
\end{equation}

Finally, we can describe the aircraft kinematics in terms of airspeed, flight path angle, and wind speed. We define the horizontal ($\dot{x}_i$) and vertical ($\dot{z}_i$) velocity components in the inertial frame and model the wind speed as a polynomial function of position in the inertial frame ($w=f(x_i,z_i)$):
\begin{align}
\dot{x}_i = v_a \cos(\gamma) + w_x \\
% \end{equation}
% \begin{equation}
\dot{z}_i = v_a sin(\gamma) + w_z
\end{align}

%%%%%%%%%%%%%%%%%%%%%%%%%%%%%%%%%%%%%%%%%%%%%%%%%%%%%%%%%%%%%%%%%%%%%%%%%%%%%%%%
%%%%%%%%%%%%%%%%%%%%%%%%%%%%%%%%%%%%%%%%%%%%%%%%%%%%%%%%%%%%%%%%%%%%%%%%%%%%%%%%
%%%%%%%%%%%%%%%%%%%%%%%%%%%%%%%%%%%%%%%%%%%%%%%%%%%%%%%%%%%%%%%%%%%%%%%%%%%%%%%%

\section{SOARING CONTROL FREEDOM}\label{soa_con_fre}

% A conventional fixed wing aircraft provides three sets of main flight control surfaces to actuate the aircraft about its principal axes. 
The ailerons, elevator, and rudder have a primary control effect on roll, pitch, and yaw respectively. Flight control in powered flight is further augmented by a throttle setpoint, which relates to the thrust reaction force. In the 6 degree of freedom (DOF) equations of motion, the aileron, elevator, rudder deflection, and thrust are the four main actuator control inputs. In this research it is useful to isolate the lateral and longitudinal motion. 

Longitudinal motion:
\begin{equation}\label{longmotion}
\begin{split}   %\usepackage{amsmath}
    & x = [\Delta u \;\; \Delta w \;\; \Delta q \;\; \Delta x \;\; \Delta z  \;\; \Delta \theta]^T\\
    & u = [\Delta \delta E \;\; \Delta \delta T]^T
\end{split}
\end{equation}

Lateral motion:
\begin{equation}
\begin{split}   %\usepackage{amsmath}
    & x = [\Delta v \;\; \Delta p \;\; \Delta r \;\; \Delta y \;\; \Delta \phi  \;\; \Delta \psi]^T\\
    &u = [\Delta \delta A \;\; \Delta \delta R]^T
\end{split}
\end{equation}

For lateral motion, the dynamics during soaring remain the same. Therefore, the continuation of this report will focus solely on the longitudinal motion. As can be seen in \autoref{longmotion}, in powered fixed-wing flight, elevator deflection and throttle setpoint are actuator inputs to the system. There are 6 state variables; position in the vertical plane, velocity in the vertical plane, pitch angle, and pitch rate. 

Consider the available control freedom in the longitudinal motion of powered fixed-wing flight. Granted that the control objectives adhere to the nonholonomic constraints of the system, one is able to satisfy 2 of the 3 DOF. Fundamentally, this allows longitudinal control systems, such as a total energy control system to function \cite{tecs}. Throttle and elevator control input are used to obtain a desired position and velocity in the vertical plane, resulting in pitch angle and pitch rate as dependent, uncontrollable variables.

% %vvvvvvvvvvvvvvvvvvvvvvvvvvvvvvvvvvv
% \iftoggle{changes}{\hl{
% ORIGINAL - CONCISENESS - STRUCTURE\\
% For lateral motion, the dynamics during soaring remain the same. Therefore, this report will focus solely on the longitudinal motion. As can be seen in \autoref{longmotion}, in powered fixed wing flight, elevator deflection and throttle setpoint are actuator inputs to the system. There are 6 state variables; position in the vertical plane, velocity in the vertical plane, pitch angle and pitch rate. 

% Consider the available control freedom in the longitudinal motion of powered fixed wing flight. Granted that the control objectives adhere to the nonholonomic constraints of the system, one is able to satisfy 2 of the 3 DOF. For example, a chosen trajectory at a chosen velocity can be achieved at the expense of pitch control freedom. This is commonly the case when steady forward flight at a given airspeed is desired. Fundamentally, this allows longitudinal control systems, such as a total energy control system to function \cite{tecs}. Throttle and elevator control input are used to obtain a desired position and velocity in the vertical plane.
% }}
% %^^^^^^^^^^^^^^^^^^^^^^^^^^^^^^^^^^^

The design of a soaring control strategy aims to eliminate throttle usage, leaving elevator deflection as the sole control actuator in the longitudinal motion. As a result, in this under-constrained system, traditional position control is not a viable option, and a novel approach is required.

% %vvvvvvvvvvvvvvvvvvvvvvvvvvvvvvvvvvv
% \iftoggle{changes}{\hl{
% ORIGINAL - CONCISENESS\\
% The objective in the design of a soaring control strategy is to eliminate any throttle usage. As a result, elevator deflection is the only control actuator in the longitudinal motion. Therefore, due to the lack of control freedom, an attempt at traditional position control in soaring flight is futile and a novel approach is required.
% }}
% %^^^^^^^^^^^^^^^^^^^^^^^^^^^^^^^^^^^

%%%%%%%%%%%%%%%%%%%%%%%%%%%%%%%%%%%%%%%%%%%%%%%%%%%%%%%%%%%%%%%%%%%%%%%%%%%%%%%%
%%%%%%%%%%%%%%%%%%%%%%%%%%%%%%%%%%%%%%%%%%%%%%%%%%%%%%%%%%%%%%%%%%%%%%%%%%%%%%%%
%%%%%%%%%%%%%%%%%%%%%%%%%%%%%%%%%%%%%%%%%%%%%%%%%%%%%%%%%%%%%%%%%%%%%%%%%%%%%%%%

\section{WIND FIELD ANALYSIS}\label{win_ana}

In the development of an orographic soaring control strategy, it is useful to consider methods to analyse and simulate a viable soaring wind field. A simplified potential flow model can be used for this, which estimates the wind field over an idealized hill with a semi-circular cross-section. \cite{Anderson,gossye2022developing}. 

% %vvvvvvvvvvvvvvvvvvvvvvvvvvvvvvvvvvv
% \iftoggle{changes}{\hl{
% ORIGINAL - CLARITY\\
% In the development of an orographic soaring control strategy, it is useful to consider methods to analyse and simulate a viable soaring wind field. 
% A simplified potential flow model for the estimation of the wind field over an idealized hill with a semi-circular cross section can be set up\cite{Anderson,gossye2022developing}. 
% }}
% %^^^^^^^^^^^^^^^^^^^^^^^^^^^^^^^^^^^

Potential flow around a cylinder can be obtained by considering a uniform stream of velocity $(U)$ and a doublet at the center of the cylinder such that the stagnation point precisely matches the boundary of the cylinder. The solution is most easily obtained in polar coordinates:
\begin{equation}
    \Phi(r,\theta) = U r (1 \text{ -- } \frac{R^2}{r^2}cos(\theta)) 
\end{equation}
The velocity in polar coordinates is then:
\begin{equation}
    V_r = \frac{\delta \Phi}{\delta r} = U (1 \text{ -- } \frac{R^2}{r^2}cos(\theta))
\end{equation}
\begin{equation}
    V_\theta = \frac{1}{r} \frac{\delta \Phi}{\delta \theta} = \text{ --}U (1 + \frac{R^2}{r^2}sin(\theta))
\end{equation}
This can be related to Cartesian coordinates by substituting $x=r cos(\theta)$ and $y=r sin(\theta)$.

\begin{figure}[ht]
\centerline{\includegraphics[width=0.9\linewidth]{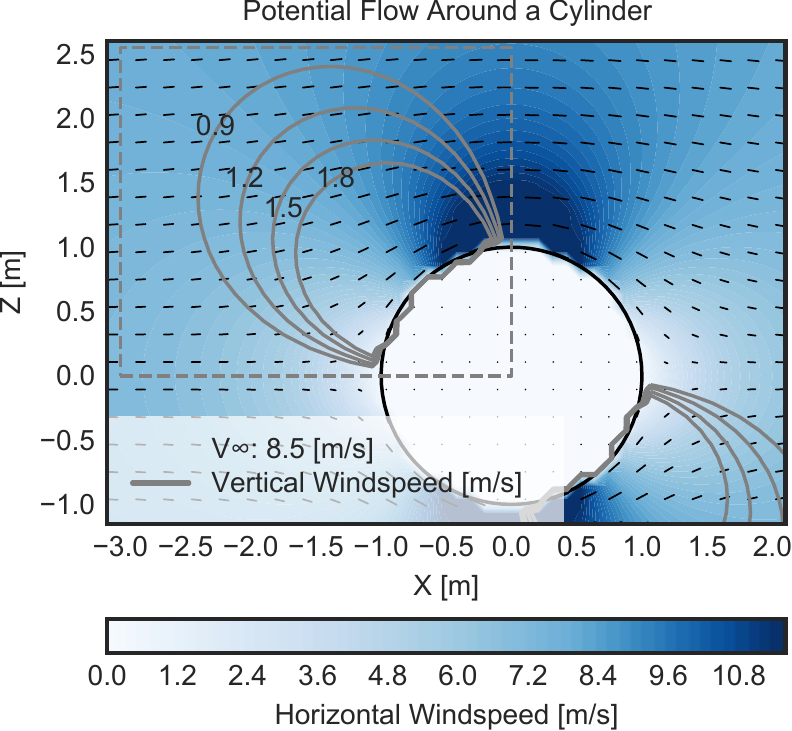}}
\caption{Wind velocity components around a cylinder obtained with potential flow theory. The upper windward quadrant is an orographic wind field.}
\label{potflow}
\end{figure}

The wind field is illustrated in \autoref{potflow}, which shows the upper windward quadrant as an orographic wind field. Of particular interest in the wind field is the vertical wind component, which should match the sink rate of the UAV to sustain soaring flight. 

% %vvvvvvvvvvvvvvvvvvvvvvvvvvvvvvvvvvv
% \iftoggle{changes}{\hl{
% ORIGINAL - CONCISENESS\\
% The resultant wind field is illustrated in \autoref{potflow}. The flow around a cylinder is of interest since the upper windward quadrant can be used as a simplified estimation of an orographic wind field. Particularly of interest in the wind field is the vertical wind component. In order to sustain soaring flight, this updraft component should match the sink rate of the UAV. 
% }}
% %^^^^^^^^^^^^^^^^^^^^^^^^^^^^^^^^^^^

Therefore, for soaring flight it is useful to consider the glide polar of an airframe, as it precisely describes the relationship between airspeed and sink rate. At the maximum endurance speed ($V_{ME}$) a particular airframe will experience the lowest rate of sink. At lower velocities the airfoil will enter its stall regime and sink rate will increase. To maintain higher velocities than $V_{ME}$ during unpowered flight, the aircraft has to assume a nose-down attitude and the sink rate will increase as well. An arbitrary quadratic function that follows the characteristics of a glide polar is chosen to study its effect in an orographic soaring wind field.

\begin{figure}[ht]
\centerline{\includegraphics[width=0.9\linewidth]{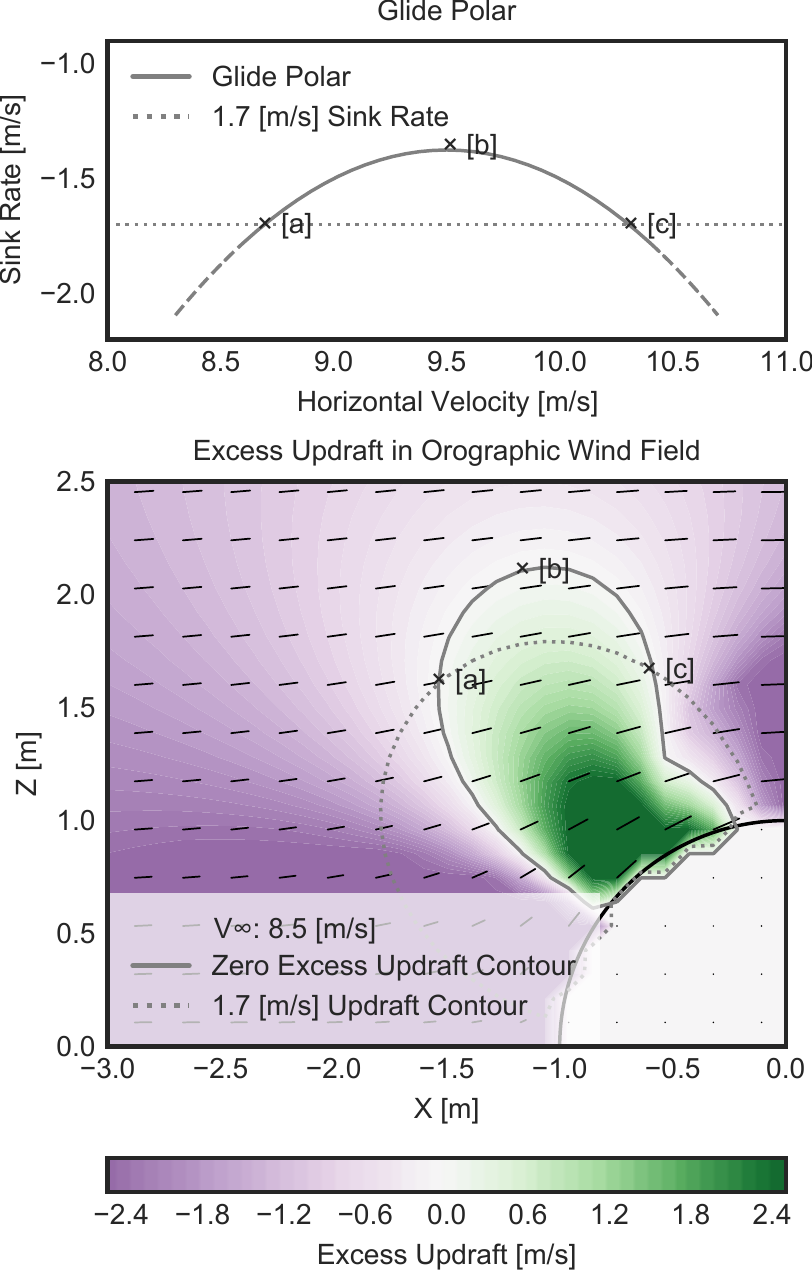}}
\caption{The glide polar defines the relation of the sink rate and horizontal velocity. Mapping of this glide polar on the orographic wind field yields the zero excess updraft contour. Three different soaring positions and their respective state on the glide polar are mapped. \vspace{-10mm}}
\label{potflowreg}
\end{figure}

% As introduced by Fisher et al. \cite{Fisher}, we can determine the feasible soaring region. At every point in the wind field, the local vertical updraft component is compared to the expected sink rate at the local horizontal wind velocity according to the glide polar. In this research we indroduce the zero excess updraft contour (ZEUC) is defined by this comparison, where the inner region shows excess updraft and the outer region a lack of sufficient updraft. The process is illustrated in \autoref{potflowreg}, where a single contour line can be seen where the local sink rate equals the local updraft. The aircraft is able to maintain its soaring position at every point on this contour. Three different soaring positions and their respective state on the glide polar are mapped. At $[a]$ and $[c]$, the aircraft will experience the same sink rate of $1.7 m/s$ at a different horizontal velocity wind component, whereas in $[b]$ the aircraft requires the least updraft to maintain its soaring position.

As introduced by Fisher et al. \cite{Fisher}, we can determine the feasible soaring region. At every point in the wind field, the local vertical updraft component is compared to the expected sink rate at the local horizontal wind velocity according to the glide polar. In this research we introduce the zero excess updraft contour (ZEUC); the line in the windfield where the expected local updraft equates the sink rate. The inner region defined by the ZEUC has an excess in updraft and the outer region has a lack of sufficient updraft. The process is illustrated in \autoref{potflowreg}. The aircraft is able to maintain its soaring position at every point on this contour. Three different soaring positions and their respective state on the glide polar are mapped. At $[a]$ and $[c]$, the aircraft will experience the same sink rate of $1.7 m/s$ at a different horizontal velocity wind component, whereas in $[b]$ the aircraft requires the least updraft to maintain its soaring position.

% %vvvvvvvvvvvvvvvvvvvvvvvvvvvvvvvvvvv
% \iftoggle{changes}{\hl{
% ORIGINAL - CLARITY\\
% As introduced by Fisher et al. \cite{Fisher}, we can determine the feasible soaring region. At every point in the wind field, the local vertical updraft component is compared to the expected sink rate at the local horizontal wind velocity according to the glide polar. Since the wind field and glide polar are continuous, a single zero excess updraft contour (ZEUC) should be expected in the orographic wind field. Defined by this contour, the inner region shows excess updraft and the outer region a lack of sufficient updraft. This process has been completed in \autoref{potflowreg}. A single contour line can be seen where the local sink rate equals the local updraft. At every point on this contour the aircraft is able to maintain its soaring position. Notice that at $[a]$ and $[c]$ the aircraft will experience the same sink rate of $1.7 m/s$ at a different horizontal velocity wind component. In $[b]$ the aircraft requires the least updraft to maintain its soaring position.
% }}
% %^^^^^^^^^^^^^^^^^^^^^^^^^^^^^^^^^^^

%%%%%%%%%%%%%%%%%%%%%%%%%%%%%%%%%%%%%%%%%%%%%%%%%

\section{AUTONOMOUS CONTROL STRATEGY}\label{aut_con_str}
As concluded in \autoref{soa_con_fre}, the longitudinal motion of a soaring UAV is an under-actuated system. The feasible region where the UAV can soar efficiently in the vertical plane is limited to a specific contour line called zero excess updraft contour (ZEUC), as outlined in \autoref{win_ana}. However, the location of this contour line cannot be determined without prior knowledge of the wind field, which complicates the use of a position controller. To address this issue, we introduce a target gradient line (TGL) as a novel approach to control the UAV's position.

% %vvvvvvvvvvvvvvvvvvvvvvvvvvvvvvvvvvv
% \iftoggle{changes}{\hl{
% ORIGINAL - CLARITY\\
% As concluded in \autoref{soa_con_fre}, the longitudinal motion of a soaring UAV is an under-actuated system. Furthermore, \autoref{win_ana} outlines how the feasible soaring region in the vertical plane for a given airframe is constrained to a single ZEUC. This further complicates the use of a position controller since the location of this contour line cannot be determined without a priori knowledge of the wind field. 
% }}
% %^^^^^^^^^^^^^^^^^^^^^^^^^^^^^^^^^^^

The TGL represents a path in the wind field along which there is a gradient in the available updraft. The TGL is chosen thoughtfully to originate at a point in the wind field where there exists excess updraft and extends upwards to a region of lack in updraft. The UAV utilizes its single degree of longitudinal control to maintain position on the TGL but is free to move along it. As a result, the UAV naturally settles in an equilibrium at the intersection of the TGL and ZEUC. This approach simplifies the control strategy, making it easier to implement and more robust to variations in the wind field.

% %vvvvvvvvvvvvvvvvvvvvvvvvvvvvvvvvvvv
% \iftoggle{changes}{\hl{
% ORIGINAL - CLARITY\\
% The presented soaring controller features a novel approach where a target gradient line (TGL) in the vertical plane is considered instead of a position setpoint. The rationale being that the UAV should utilize its single degree of longitudinal control to maintain position on this line, but is otherwise free to move along this line. A thoughtful choice of TGL originates at a point in the wind field where there exists excess updraft. The line should extent upwards to a region of lack in updraft. As a result, the UAV will naturally settle in an equilibrium at the intersection of the TGL and ZEUC.
% }}
% %^^^^^^^^^^^^^^^^^^^^^^^^^^^^^^^^^^^

The natural equilibrium soaring location of the UAV can be influenced by several factors. Firstly, the operator can manipulate the UAV's position along the ZEUC by rotating or translating the TGL in the vertical plane, thereby realizing a single degree of control freedom. Secondly, during flight, small changes in the wind field are expected, which can result in a changing position of the ZEUC. By maintaining position on the TGL, The UAV will naturally move along the TGL to a new equilibrium point, which will be at the intersection with the ZEUC. 

% %vvvvvvvvvvvvvvvvvvvvvvvvvvvvvvvvvvv
% \iftoggle{changes}{\hl{
% ORIGINAL - CLARITY - ACCURACY\\
% Several deliberate factors and external factors can influence the natural equilibrium point.
% Firstly, by the operator rotating or translating the TGL in the vertical plane, the UAV can be manipulated to move along the ZEUC. As such, a single degree of control freedom can be realized. Secondly, during flight, small changes in the wind field should be expected. This can be as a result of changing wind speed or otherwise changing flow deflection. These result in a changing position of the ZEUC, which the controller accommodates by converging on a new equilibrium point along the chosen TGL. 
% }}
% %^^^^^^^^^^^^^^^^^^^^^^^^^^^^^^^^^^^

\begin{figure}[]
\centerline{\includegraphics[width=0.9\linewidth]{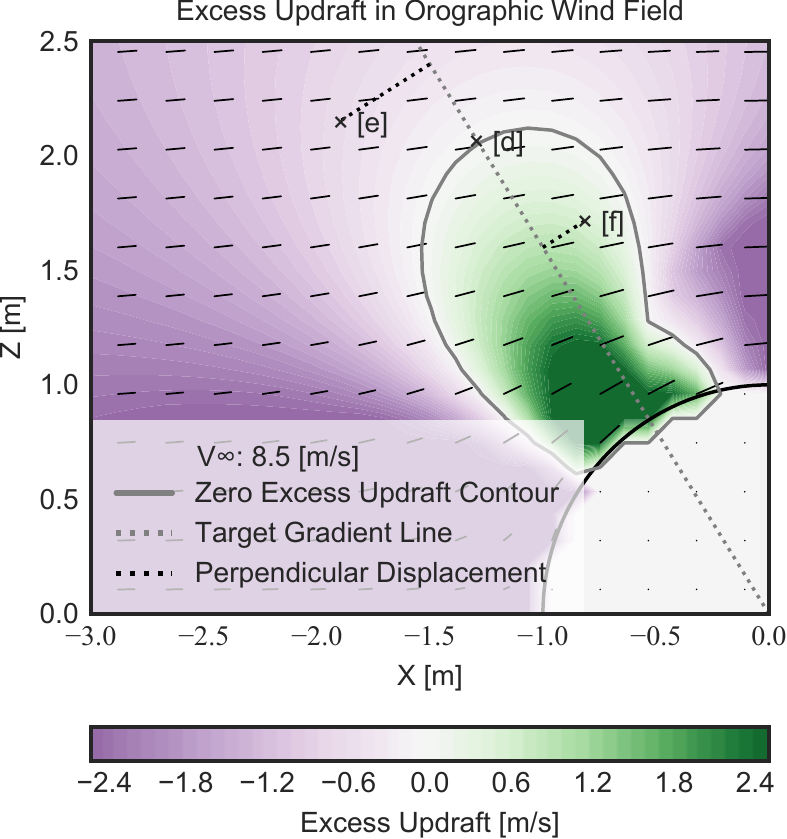}}
\caption{The autonomous control strategy considers a target gradient line intersecting the zero excess updraft contour. The controller response is proportional to the perpendicular displacement to the TGL.\vspace{-6mm}}
\label{B_pot_tgl}
\end{figure}

The controller to maintain position on the TGL is implemented as a closed-loop pitch controller. The perpendicular distance of the UAV to the TGL, $(e_{\rho})$, is formulated as an error input for the controller. By convention, $e_{\rho}$ is positive when the UAV is upstream of the TGL and negative when the UAV is downstream, defined in \autoref{eq:e_rho}.

\begin{equation}
\label{eq:e_rho}
    e_{\rho} = s * \frac{\mid Ax_1 + Bz_1 + C \mid}{A^2 + B^2}  \quad \text{with }    \quad
    \begin{aligned}
        s &= \hspace{0.76em} 1  \text{ IF upstream} \\
        s &=                -1 \text{ IF downstream}    
    \end{aligned}
\end{equation}

With $TGL: Ax+Bz+C = 0$ and UAV position $P:(x_1,z_1)$. The implementation is the test setup is analogous, where instead the TGL is extruded along $y$, and the perpendicular distance to said target plane is considered.

The pitch setpoint is then obtained as follows:
\begin{equation}
\label{eq:thetasp}
\theta_{sp} = \theta_{0} + k_p e_{\rho} +  k_i \int e_{\rho} dt + k_d \frac{de_{\rho}}{dt}
\end{equation}

$\theta_{0}$ is the trimmed pitch angle at the expected flight velocity. Stable soaring flight can be achieved by tuning the proportional ($k_p$) and derivative ($k_d$) gains. The use of an integral gain ($k_i$) is recommended to minimise steady-state error and realise full convergence to the TGL. The elevator setpoint $(e_{sp})$ is controlled with a closed-loop controller, taking as input the pitch error $(\theta_e)$.

\begin{equation}
\label{eq:a_sp}
e_{sp} = k_p \theta_e +  k_i \int \theta_e dt + k_d \frac{d\theta_e}{dt}
\end{equation}

The TGL can be thoughtfully chosen to best deal with disturbances to the equilibrium. Namely, the total energy state of the vehicle in immediate proximity to the TGL should be considered. For instance, a horizontal TGL would often be a poor choice. A vehicle that finds itself below the TGL might lack the potential energy as well as higher updraft regions to regain altitude towards its TGL. Furthermore, a TGL is best chosen roughly perpendicular to the ZEUC. This way, minimal displacement along the TGL is required to accommodate changes in the wind field. It is important to note that this control strategy does not require a priori knowledge of the wind field. However, a general estimate of the shape of the wind field, such as knowledge of the location of the updraft core, is desired to effectively choose a TGL.

%%%%%%%%%%%%%%%%%%%%%%%%%%%%%%%%%%%%%%%%%%%%%%%%%%%%%%%%%%%%%%%%%%%%%%%%%%%%%%%%
%%%%%%%%%%%%%%%%%%%%%%%%%%%%%%%%%%%%%%%%%%%%%%%%%%%%%%%%%%%%%%%%%%%%%%%%%%%%%%%%
%%%%%%%%%%%%%%%%%%%%%%%%%%%%%%%%%%%%%%%%%%%%%%%%%%%%%%%%%%%%%%%%%%%%%%%%%%%%%%%%

\section{EXPERIMENTAL TEST SETUP}\label{exp_tes_set}
\begin{figure}[ht]
\centerline{\includegraphics[width=0.9\linewidth]{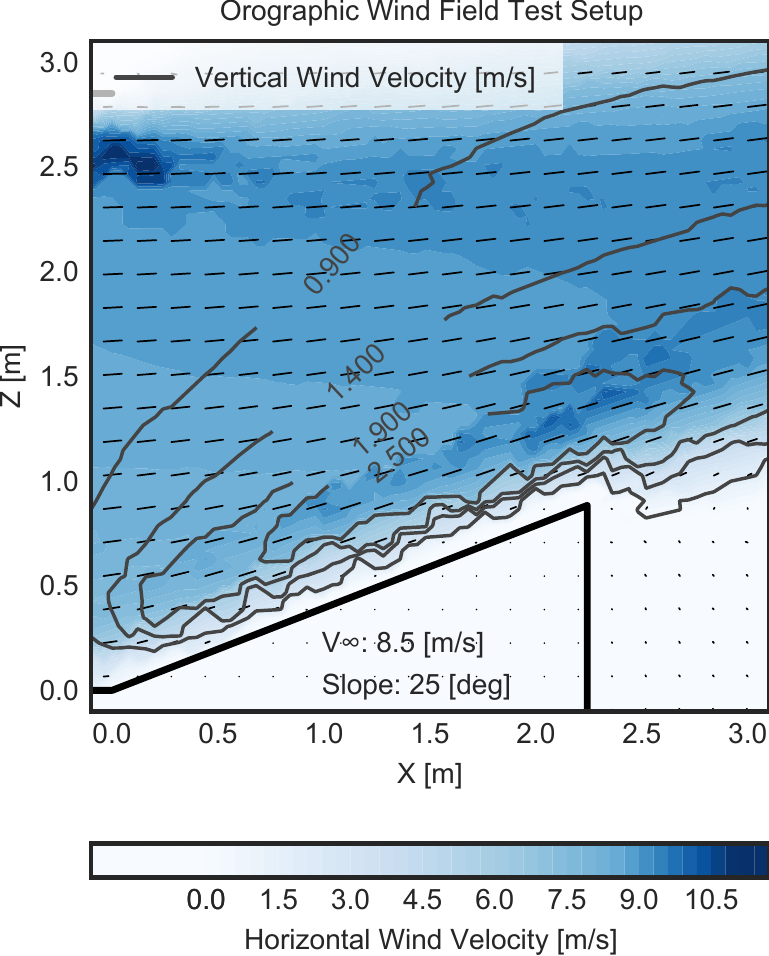}}
\caption{Wind velocity components of the orographic wind field in the experimental test setup, obtained by CFD. }
\label{testwinfvel}
\end{figure}

We conducted a full-scale test campaign in the open jet facility at Delft University of Technology. The facility has an outlet cross-section of $2.85 m$ and can generate wind velocities up to $35 m/s$. An updraft was created by placing a board at various angles in the airflow. We used this geometry to create a CFD model to estimate wind velocity components in the test section at various wind velocities and slope positions. \cite{ANSYS}. The geometry is highlighted in \autoref{intropic} and the velocity components are shown in \autoref{testwinfvel}.

% %vvvvvvvvvvvvvvvvvvvvvvvvvvvvvvvvvvv
% \iftoggle{changes}{\hl{
% ORIGINAL - CONCISENESS\\A full scale test campaign was conducted in the open jet facility (OJF) at Delft University of Technology. This wind tunnel with an outlet cross section of $2.85 m$ is able to precisely generate a wind velocity of up to $35 m/s$. An updraft was created by placing a board directly in the airflow at various angles. Using this geometry, a CFD model was created to estimate the wind velocity components in the test section \cite{ANSYS}. Although three-dimensional simulations were conducted at various wind velocities and slope positions, only the local wind components on the centered vertical plane are used to gain insight in the longitudinal motion of the test aircraft. The geometry is highlighted in \autoref{intropic} and the velocity components are shown in \autoref{testwinfvel}.
% }}
% %^^^^^^^^^^^^^^^^^^^^^^^^^^^^^^^^^^^

\begin{figure}[ht]
\centerline{\includegraphics[width=0.95\linewidth]{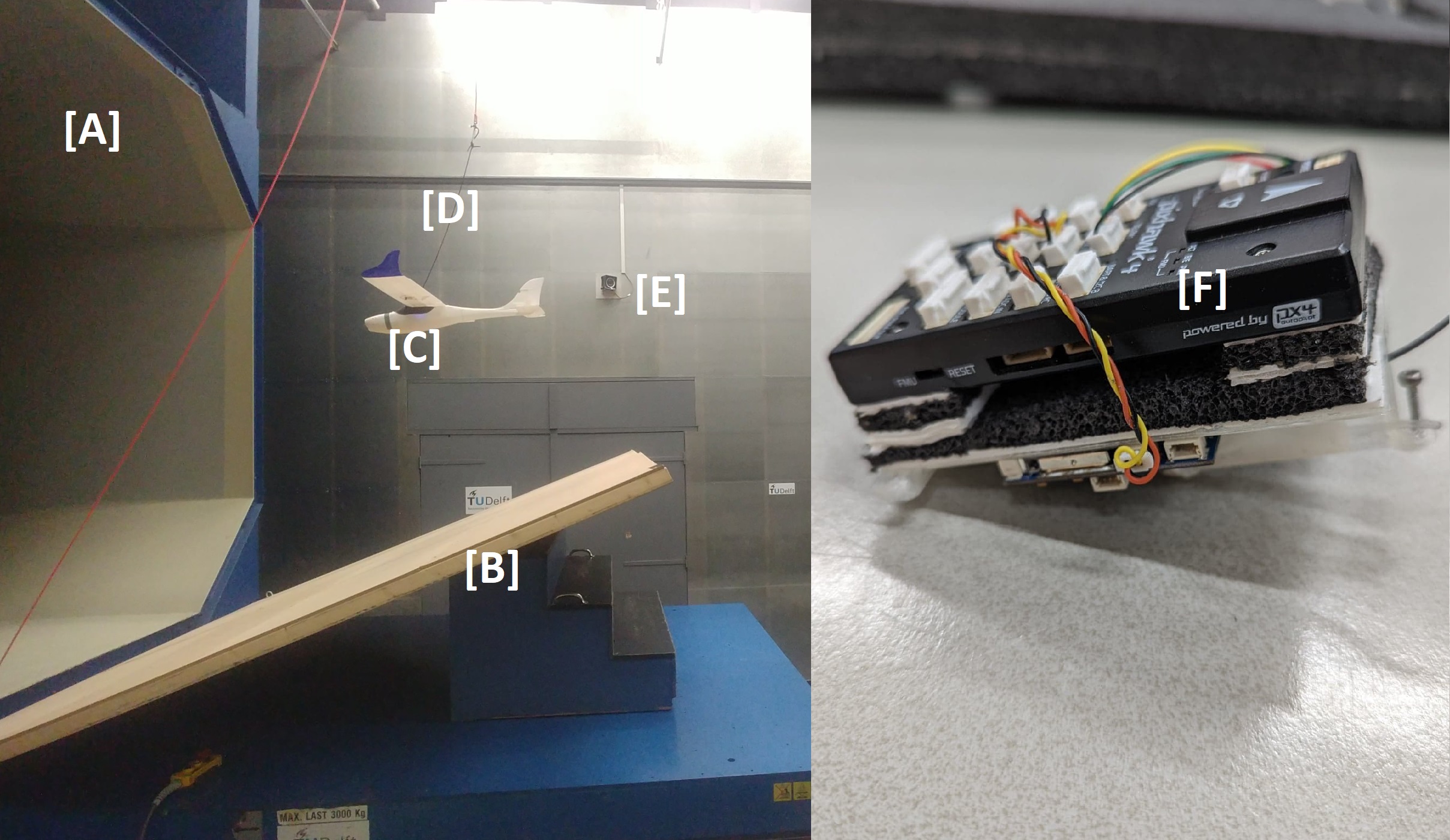}}
\caption{Test setup and data collection. [A] Open Jet Facility wind tunnel, [B] Adjustable slope, [C] UAV (without propeller mounted), [D] Safety tether, [E] Optitrack system camera, [F] Pixhawk 4 running Paparazzi autopilot}
\label{teststup}
\end{figure}

The test UAV was a modified Eclipson Model C \cite{Eclipson} 3D-printed model aircraft running Paparazzi autopilot \cite{Paparazzi}. The aircraft had three degrees of actuation with aileron, rudder, and elevator but no propeller. We determined the aircraft's glide polar by third-order polynomial regression of gliding flight data at discretely different pitch attitudes. This glide polar is shown in \autoref{tglcontr}.
We used an Optitrack system \cite{Optitrack}, mounted in the test facility, to receive the aircraft's positioning data, which was also logged to evaluate the controller's performance. An image of the test setup and its components is shown in \autoref{teststup}.

% %vvvvvvvvvvvvvvvvvvvvvvvvvvvvvvvvvvv
% \iftoggle{changes}{\hl{
% ORIGINAL - CONCISENESS\\
% The test UAV was a modified Eclipson Model C \cite{Eclipson} 3D printed model aircraft running paparazzi \cite{Paparazzi} autopilot. There is no propeller present and the aircraft has three degrees of actuation in a traditional setup with aileron, rudder and elevator. The glide polar of the aircraft used in this research was determined by third order polynomial regression of gliding flight data at discretely different pitch attitudes. This glide polar is shown in \autoref{tglcontr}.
% }}
% %^^^^^^^^^^^^^^^^^^^^^^^^^^^^^^^^^^^

% \begin{figure}[ht]
% \centerline{\includegraphics[width=0.85\linewidth]{figures/C6-polar.pdf}}
% \caption{Eclipson Model C glide polar obtained by polynomial regression of flight data points.}
% \label{exp_glidepolar}
% \end{figure}

% %vvvvvvvvvvvvvvvvvvvvvvvvvvvvvvvvvvv
% \iftoggle{changes}{\hl{
% ORIGINAL - CONCISENESS\\
% The aircraft receives its positioning data from an Optitrack system \cite{Optitrack}, mounted in the test facility. This positioning data is also logged to evaluate the performance of the controller. An image of the test setup and its components is shown in \autoref{teststup}.
% }}
% %^^^^^^^^^^^^^^^^^^^^^^^^^^^^^^^^^^^

To ensure lateral stability and heading during testing, we implemented two lateral closed-loop control systems. The roll controller affects the ailerons and keeps the aircraft level, while the yaw controller affects the rudder and maintains the heading towards a virtual waypoint located $5 m$ upwind from the wind tunnel settling chamber, achieving centering within the wind tunnel cross-section. The yaw error is defined as $e_{\psi} = \tan(y/R) - \psi$, where $y$ is the displacement from the vertical center plane in the wind tunnel and $R$ is the distance to the virtual upwind waypoint.

Both actuator setpoints are obtained through a closed-loop control system with proportional ($k_p$), integral ($k_i$), and derivative ($k_d$) gains, as shown in Equations \ref{eq:a_sp} and \ref{eq:r_sp}. The novel soaring controller, presented in \autoref{aut_con_str}, affects the elevator.

\begin{equation}
\label{eq:a_sp}
a_{sp} = k_p e_{\phi} +  k_i \int e_{\phi} dt + k_d \frac{de_{\phi}}{dt}
\end{equation}

\begin{equation}
\label{eq:r_sp}
r_{sp} = k_p e_{\psi} + k_d \frac{de_{\psi}}{dt}
\end{equation}

Our primary goal was to validate the novel soaring controller and investigate the effect of changing the slope, wind speed, and placement of the TGL.

% %vvvvvvvvvvvvvvvvvvvvvvvvvvvvvvvvvvv
% \iftoggle{changes}{\hl{
% ORIGINAL - CONTENT\\
% In order to maintain lateral stability and heading during testing, two lateral control loops were implemented. A roll controller was tasked with keeping the aircraft level using its set of ailerons. A yaw controller effecting the rudder was tasked with maintaining heading towards a virtual waypoint $5 m$ upwind from the wind tunnel settling chamber, realizing centering within the wind tunnel cross section. The novel soaring controller effects the elevator and is responsible for its longitudinal motion.

% The primary goal of the testing campaign was to validate the novel soaring controller. Furthermore, the effect of changing the slope, the wind speed and the placement of the TGL were investigated.
% }}
% %^^^^^^^^^^^^^^^^^^^^^^^^^^^^^^^^^^^

%%%%%%%%%%%%%%%%%%%%%%%%%%%%%%%%%%%%%%%%%%%%%%%%%%%%%%%%%%%%%%%%%%%%%%%%%%%%%%%%
%%%%%%%%%%%%%%%%%%%%%%%%%%%%%%%%%%%%%%%%%%%%%%%%%%%%%%%%%%%%%%%%%%%%%%%%%%%%%%%%
%%%%%%%%%%%%%%%%%%%%%%%%%%%%%%%%%%%%%%%%%%%%%%%%%%%%%%%%%%%%%%%%%%%%%%%%%%%%%%%%

\section{TEST RESULTS AND DISCUSSION}\label{Test_res}
\begin{figure}[ht]
\centerline{\includegraphics[width=0.88\linewidth]{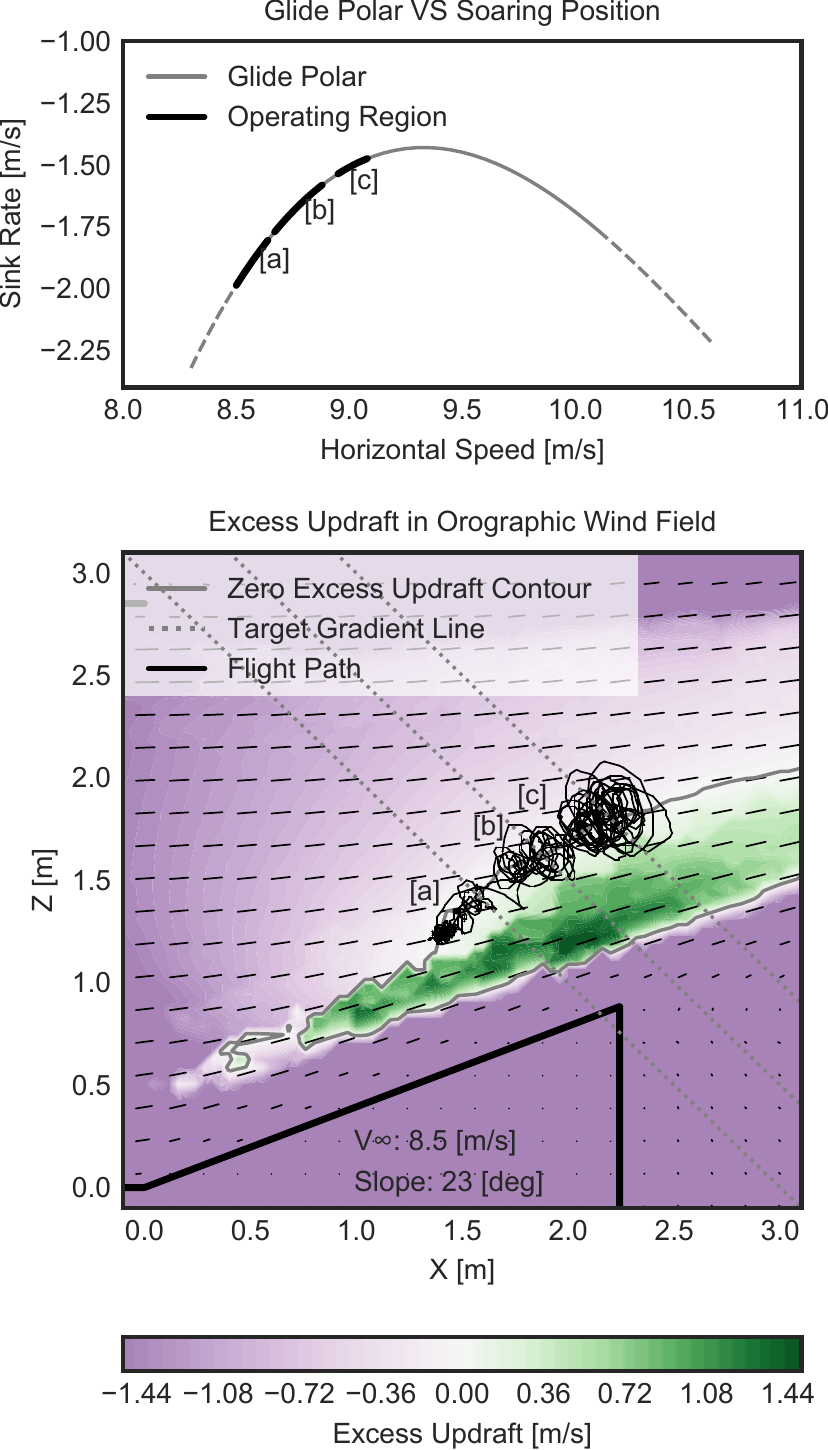}}
\caption{Mapping of the Eclipson Model C glide polar on the orographic wind field of the experimental test setup, yielding the expected zero excess updraft contour. Three different TGL positions are tested and the corresponding flight path is plotted. Note that the flight path corresponds well with the expected soaring locations. \vspace{-6mm}}
\label{tglcontr}
\end{figure}

By combining the velocity components in the wind field from CFD simulations and the glide polar, the (ZEUC) can be generated, as shown in \autoref{tglcontr}. It should be noted that the required updraft along the ZEUC is not a constant amount. It is a function of the local horizontal wind velocity component and the glide polar.

% %vvvvvvvvvvvvvvvvvvvvvvvvvvvvvvvvvvv
% \iftoggle{changes}{\hl{
% ORIGINAL - STYLE\\
% Combining the velocity components in the wind field from CFD simulation and the glide polar, allows for the generation of the ZEUC line, as seen in \autoref{tglcontr}. Note that the required updraft along the ZEUC is not a constant amount. It is a function of the local horizontal wind velocity component and the glide polar. % obtained in \autoref{exp_glidepolar}.
% }}
% %^^^^^^^^^^^^^^^^^^^^^^^^^^^^^^^^^^^

In \autoref{tglcontr}, consider a test with the leftmost static TGL at position $[a]$. After manual tuning of the controller gains, it is observed from the flight path that the controller is able to successfully maintain position with minimal oscillations. Note that the TGL was chosen to be roughly perpendicular to the ZEUC. Defining a TGL less perpendicular to the ZEUC negatively affected the controller performance with larger oscillations and drift from the TGL.

As we translate the TGL to positions $[b]$ and $[c]$, the settled equilibrium of the aircraft also moves along with the TGL. Each equilibrium point corresponds to a different part of the aircraft's glide polar. Notably, the flight path positions recorded during the test closely coincide with the intersection of the TGL and ZEUC. This confirms the effectiveness of the controller in maintaining position on the chosen TGL and the accuracy of the estimated wind field and glide polar. Additionally, moving the TGL proves to be an effective method for achieving a single degree of position control freedom with this soaring controller. Larger oscillations were observed downstream as a result of overshoot due to the increased elevator effectiveness at higher airspeed.

% %vvvvvvvvvvvvvvvvvvvvvvvvvvvvvvvvvvv
% \iftoggle{changes}{\hl{
% ORIGINAL - CONCISENESS - CONTENT \\
% In \autoref{tglcontr}, consider a test with the left most static TGL at position $[a]$. It is observed that the controlled is able to successfully maintain position with minimal oscillations. It is now interesting to study the effect of incrementally translating the TGL during flight. In position $[b]$ and $[c]$ the settled equilibrium of the aircraft moves along with the translation of the TGL. Also note that aircraft will operates in a different part of its glide polar at every equilibrium point. Furthermore, it is pleasing to see that the recorded flight path segment positions coincides almost perfectly with the intersection of the TGL and ZEUC. Firstly, this highlights that the controller is effective at maintaining position on the chosen TGL. Secondly, the calculated ZEUC, resulting from the estimated wind field and estimated glide polar corresponds closely to the actual soaring region recorded during the flight test. This test also illustrates that moving the TGL is an effective method of realizing a single degree of position control freedom with this soaring controller. 
% }}
% %^^^^^^^^^^^^^^^^^^^^^^^^^^^^^^^^^^^

\begin{figure}[ht]
\centerline{\includegraphics[width=0.9\linewidth]{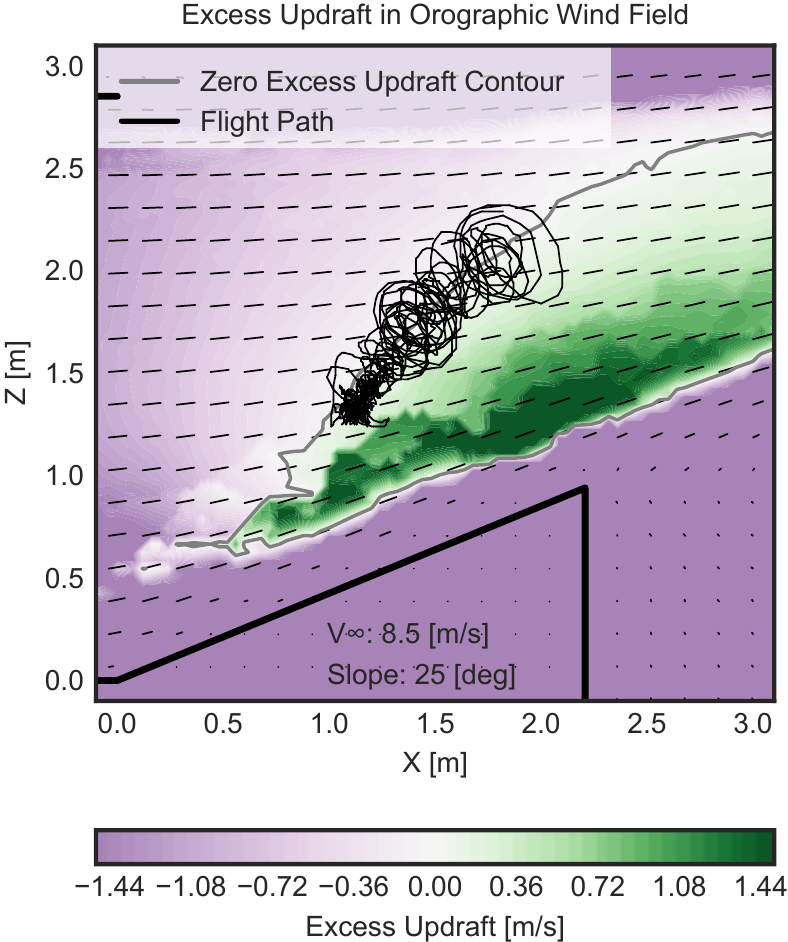}}
\caption{Effect of changing the slope from $23 deg$ to $25 deg$. The zero excess updraft contour (ZEUC) is now located noticeably higher. The flight path is plotted over a translating TGL. From the flight track we observe that the soaring location corresponds with the newly obtained ZEUC. \vspace{-2mm}}
\label{slpcontr}
\end{figure}

In \autoref{slpcontr}, we investigated the effect of changing the slope in the test setup on the resultant ZEUC. We increased the slope from $23$ to $25$ degrees and observed that the contour shifted upwards. To study the controller's adaptability to changes in the wind field, we repeated the testing with incremental translation of the TGL at this higher slope. Our results show that the controller was able to adapt to changes in the wind field while maintaining the same level of control freedom.

% %vvvvvvvvvvvvvvvvvvvvvvvvvvvvvvvvvvv
% \iftoggle{changes}{\hl{
% ORIGINAL - CONCISENESS \\
% Consider in \autoref{slpcontr}, the effect of changing the slope in the test setup on the resultant ZEUC. In comparison with \autoref{tglcontr}, the slope was increased from $23 deg$ to $25 deg$. Note how this causes the contour to shift upwards. Testing with incremental translation of the TGL was repeated at this higher slope. It can be seen that the controller is able to adapt to changes in the wind field while maintaining the same level of control freedom. 
% }}
% %^^^^^^^^^^^^^^^^^^^^^^^^^^^^^^^^^^^

Next, we examined the effect of changes in wind speed on the UAV's performance. We incrementally increased the wind speed from $8.5$ to $9.5$ m/s while maintaining the same TGL. Throughout this range, the UAV was able to successfully maintain soaring flight, and we did not observe any significant changes in its hovering position. This unexpected result can be explained by considering the immediate effect of changes in horizontal wind velocity on the updraft and sink rate. We assumed that the vertical updraft component would scale proportionally to the horizontal wind. Therefore, a change in wind velocity would not change the shape of the wind field, but it would only scale the magnitude of all local wind vectors.

% %vvvvvvvvvvvvvvvvvvvvvvvvvvvvvvvvvvv
% \iftoggle{changes}{\hl{
% ORIGINAL - CONCISENESS - STYLE \\
% Finally, the effect of changes in wind speed can be studied. In the performed test the wind speed was incrementally increased from $8.5 m/s$ to $9.5 m/s$, while maintaining the same TGL. Throughout this range, the UAV was able to successfully maintain soaring flight. Surprisingly no significant change in hovering position could be noted. However, the results can be explained by considering the immediate effect of changes in horizontal wind velocity on the updraft and sink rate. It is assumed that the vertical updraft component would scale proportionally to the horizontal wind. Effectively this means that a change in wind velocity will not change the shape of the wind field but rather just scale the magnitude of all local wind vectors. 
% }}
% %^^^^^^^^^^^^^^^^^^^^^^^^^^^^^^^^^^^

\begin{figure}[ht]
\centerline{\includegraphics[width=0.9\linewidth]{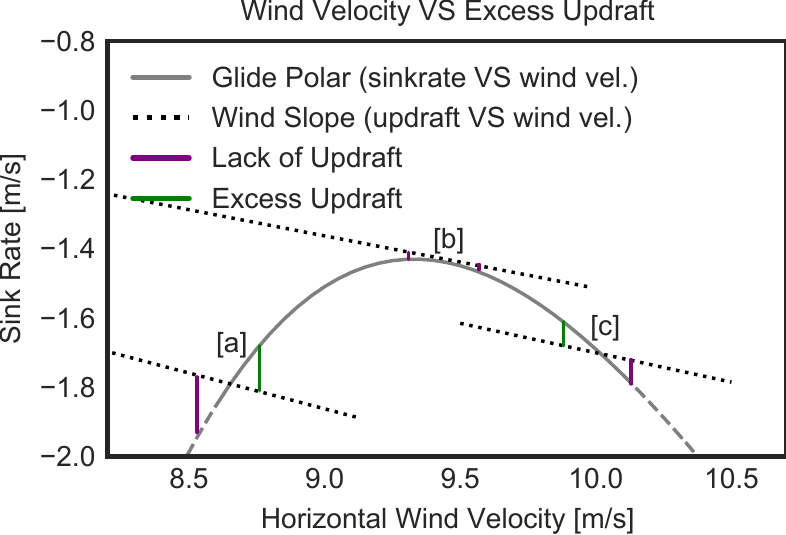}} 
\caption{Comparison of the immediate change in sink rate and updraft as a result of changing the horizontal wind velocity at different segments on the glide slope. Changes in wind velocity affect the generated updraft in the wind field and effective sink rate of the aircraft differently.}
\label{C9-vel}
\end{figure}

Consider \autoref{C9-vel}, which presents three scenarios where the updraft and sink rate are initially balanced.
\begin{itemize}
\item In scenario [a], the airfoil is in the stall regime and a change in wind velocity has a significant impact on the updraft and sink rate. An increase in wind velocity leads to an increase in the updraft component and a decrease in the sink rate, resulting in a net upward movement. Conversely, a decrease in wind velocity causes a net downward movement.

\item In scenario [c], a change in wind velocity has a greater impact on the sink rate than the updraft, leading to a net downward movement with an increase in wind velocity and a net upward movement with a decrease in wind velocity.

\item Finally, when the aircraft is operating near its optimal glide speed in the vicinity of [b], changes in wind velocity cause both the updraft and sink rate to change at a comparable rate, resulting in minimal movement. This scenario was observed during the experimental test and helps explain that limited movement was observed.
\end{itemize}

%vvvvvvvvvvvvvvvvvvvvvvvvvvvvvvvvvvv
% \iftoggle{changes}{\hl{
% ORIGINAL - CONCISENESS - STYLE \\
% Consider \autoref{C9-vel}, highlighting three scenarios where updraft and sink rate are initially balanced. Observing the effect in $[a]$, when operating in the stall regime of the airfoil, an increase in wind velocity will increase the updraft component and decrease the sink rate, causing a net upwards movement. Analogously, a decrease in wind velocity causes a net downward movement. Considering point $[c]$, an increase in wind velocity will increase the sink rate at a higher rate than the updraft. In this case, a downwards movement will result. By similar fashion we see that a decrease in wind velocity causes an upwards movement. When the aircraft is operating in the proximity of its optimal glide speed, in the vicinity of $[b]$, both the updraft and sink rate change at a comparable rate as a result of changes in the wind velocity. This was the case during the experimental test and would explain why little movement was observed. 
% }}
%^^^^^^^^^^^^^^^^^^^^^^^^^^^^^^^^^^^
A change in wind velocity alters the shape and position of the ZEUC accordingly. The reaction force resulting from an imbalance between updraft and sink rate allows the aircraft to settle on the newly obtained ZEUC. 

%%%%%%%%%%%%%%%%%%%%%%%%%%%%%%%%%%%%%%%%%%%%%%%%%

\section{CONCLUSIONS}

The objective of this research was to demonstrate the feasibility of autonomous orographic soaring for fixed-wing UAVs. We identified the feasible soaring region, which can be represented by a single line known as the zero excess updraft contour (ZEUC).

As the longitudinal motion of a soaring UAV is an under-actuated system, we introduced the concept of a target gradient line (TGL) to provide a single degree of control freedom. We then presented an autonomous controller that enables position keeping at the intersection of the TGL and ZEUC. We validated the controller in an experimental test setup, and the results showed that it effectively maintained position on the chosen TGL without using any thrust as the UAV had no propeller. Furthermore, the position of the logged flight segments closely aligned with the expected ZEUC, which was derived from the estimated wind field and glide polar.

We demonstrated that adjusting the TGL is an effective way to realize a single degree of control freedom in the system. Finally, we showed that the controller is robust to changes in the wind field, such as alterations in slope or changes in the free-stream velocity of the wind tunnel.

% %vvvvvvvvvvvvvvvvvvvvvvvvvvvvvvvvvvv
% \iftoggle{changes}{\hl{
% ORIGINAL - CONCISENESS - STYLE \\
% The aim of this research was to demonstrate autonomous orographic soaring for fixed wing UAVs. In order to gain insight in an orographic soaring field, a potential flow model around a cylinder was set up. An approximated glide polar was mapped on this wind field in order to determine the feasible soaring region. In a continuous wind field this feasible region is condensed to a single line, formerly called the zero excess updraft contour (ZEUC). The longitudinal motion of a soaring UAV is an under-actuated system. This research introduced the concept of an arbitrary target gradient line (TGL) to realise its single degree of control freedom. An autonomous controller was presented that enables position keeping on the intersection of aforementioned two lines. This controller was then validated in an experimental test setup. Test results show that the controller is effective at maintaining position on the chosen TGL. Furthermore, the position of the logged flight segments lie on the expected ZEUC that resulted from combining the estimated wind field and estimated glide polar. It was demonstrated that moving the TGL is an effective way of realizing a single degree of control freedom in the system. Finally, the controller was shown to be robust in the case of a changing wind field, by means of altering the slope, and changing the free-stream velocity of the wind tunnel. 
% }}
% %^^^^^^^^^^^^^^^^^^^^^^^^^^^^^^^^^^^
The performed tests in this research were limited by the cross-section of the wind tunnel. To enable a larger orographic wind field and more diverse wind conditions, additional testing in an outdoor environment is recommended. Furthermore, this would enable testing in a wider envelope of the UAV's glide polar. When soaring in a broad airspeed range, it is recommended to adjust the gains for changes in elevator effectiveness. Finally, it can be challenging to set a favorable TGL without a priori knowledge of the wind field. Further research on obtaining an initial soaring position is suggested.

\bibliographystyle{IEEEtran}
\bibliography{IEEEabrv,IEEEexample.bib}

\end{document}